\documentclass[doublecolumn]{IEEEtran}

\usepackage{graphicx}
\usepackage{subfig}
\usepackage{amsmath}
\usepackage{amssymb}
\usepackage[font=footnotesize]{caption} 
\usepackage{subfig} 
\usepackage[noadjust]{cite} 
\usepackage{color}
\usepackage{algorithm} 
\usepackage{algpseudocode} 
\usepackage{enumerate}
\usepackage[hidelinks,colorlinks=false]{hyperref}
\usepackage[left=0.75in, right=0.75in, top=0.75in, bottom=0.75in]{geometry}
\usepackage{authblk} 
\usepackage{arydshln} 

\usepackage{bm}
\usepackage{tikz}
\usetikzlibrary{calc} 
\usetikzlibrary{shapes} 
\usetikzlibrary{chains}
\usetikzlibrary{fit}
\usetikzlibrary{arrows}
\usetikzlibrary{decorations.text} 
\usetikzlibrary{decorations.markings}
\usetikzlibrary{decorations.pathmorphing} 
\usetikzlibrary{shadows}
\usetikzlibrary{patterns}
\usetikzlibrary{matrix}
\usepackage{pgfplots}
\usepackage[europeanresistors]{circuitikz}
\usepackage[outline]{contour} 
\contourlength{1.5pt}
\usetikzlibrary{fadings}
\usetikzlibrary{patterns}
\usetikzlibrary{shapes}

\newtheorem{lemma}{Lemma}

\graphicspath{{figures/}}

\begin{document}
	
\title{Motion Planning for Aerial Pick-and-Place based on Geometric Feasibility Constraints}
\author{Huazi~Cao, Jiahao~Shen, Cunjia~Liu, Bo~Zhu, Shiyu~Zhao
	\thanks{H. Cao, J. Shen, and S. Zhao are with the School of Engineering at Westlake University, Hangzhou, China. {\tt\small \{caohuazi,shenjiahao,zhaoshiyu\}@westlake.edu.cn}}
	\thanks{C. Liu is with the Department of Aeronautical and Automotive Engineering at Loughborough University, Loughborough, UK. {\tt\small c.liu5@lboro.ac.uk}}
	\thanks{B. Zhu is with the School of Aeronautics and Astronautics at Sun Yat-sen University, Guangzhou, China. {\tt\small zhubo5@mail.sysu.edu.cn}}
}
	\maketitle
	\begin{abstract}
		This paper studies the motion planning problem of the pick-and-place of an aerial manipulator that consists of a quadcopter flying base and a Delta arm. We propose a novel partially decoupled motion planning framework to solve this problem. Compared to the state-of-the-art approaches, the proposed one has two novel features. First, it does not suffer from increased computation in high-dimensional configuration spaces. That is because it calculates the trajectories of the quadcopter base and the end-effector separately in the Cartesian space based on proposed geometric feasibility constraints. The geometric feasibility constraints can ensure the resulting trajectories satisfy the aerial manipulator's geometry. Second, collision avoidance for the Delta arm is achieved through an iterative approach based on a pinhole mapping method, so that the feasible trajectory can be found in an efficient manner. The proposed approach is verified by three experiments on a real aerial manipulation platform. The experimental results show the effectiveness of the proposed method for the aerial pick-and-place task. 
	\end{abstract}

\def\abstractname{Note to Practitioners}
\begin{abstract}
Aerial manipulators have attracted increasing research interest in recent years due to their potential applications in various domains. In this paper, we particularly focus on the motion planning problem of the pick-and-place of aerial manipulators. We propose a novel partially decoupled motion planning framework, which calculates the trajectories of the quadcopter base and the end-effector in Cartesian space, respectively. Geometric feasibility constraints are proposed to coordinate the trajectories to ensure successful execution. Three experiments on a real aerial manipulator platform demonstrate the effectiveness of the approach. In future research, we will address the motion planning problem of aerial manipulators in complex environments.
\end{abstract}
	
	
\section{Introduction}

An aerial manipulator is a novel type of flying robot that consists of a multirotor and a robotic arm. Due to their ability to move quickly and operate precisely in high-altitude and complex workspaces, the aerial manipulator has potential applications in various domains, including transportation, inspection, and maintenance (see \cite{ollero2021past, xilun2019review, mohiuddin2020survey, ruggiero2018aerial} for recent surveys).

\begin{figure}[t]
	\centering
	\includegraphics[width=1\linewidth]{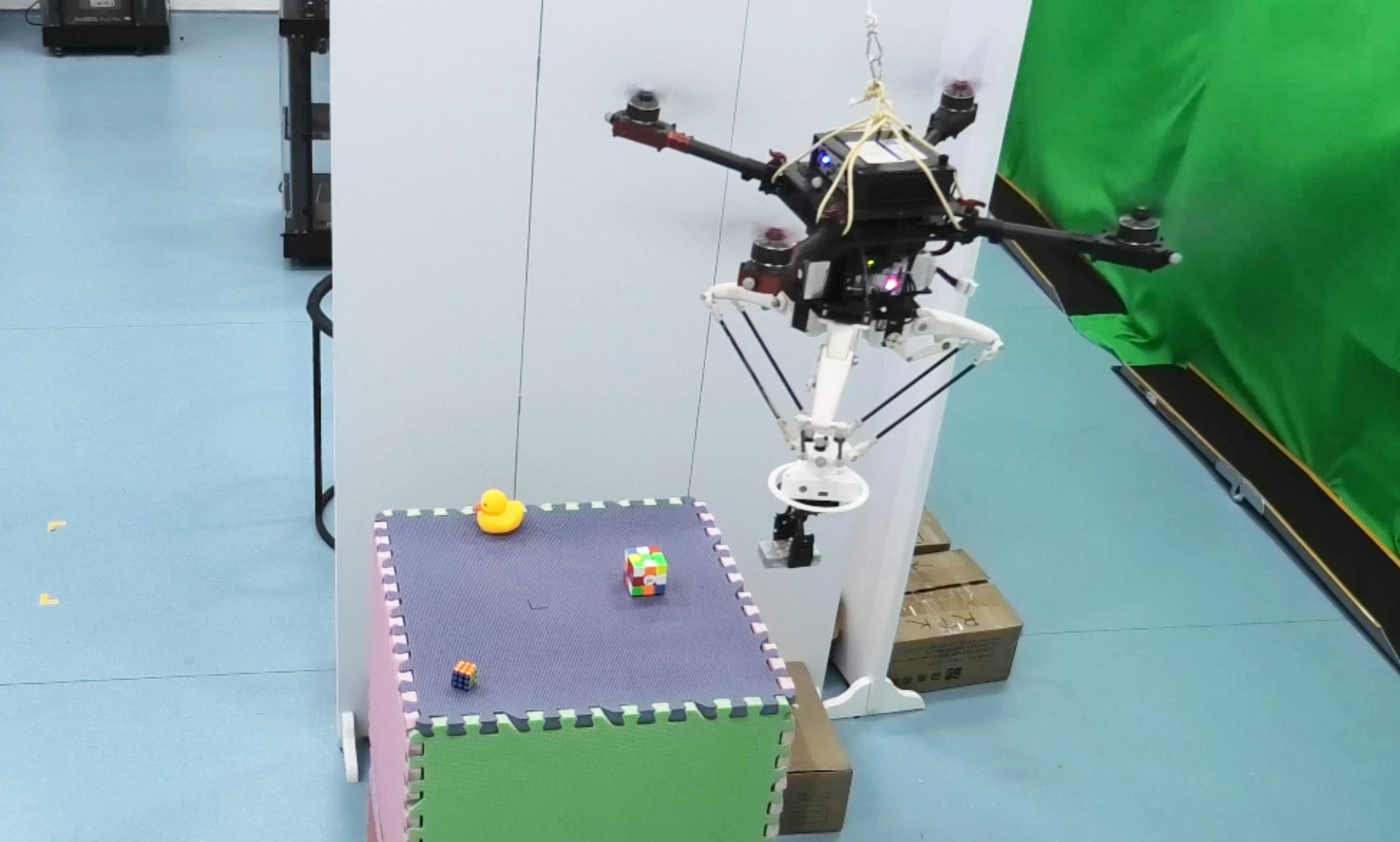}
	\caption{Aerial pick-and-place by an aerial manipulator. The experimental video is available at https://youtu.be/q7O9v7l2Oho.}
	\label{fig_result_aerial_pick_place}
\end{figure}

Aerial manipulation has been studied from various aspects such as platform design \cite{bodie2020active,zhang2022aerial, nguyen2018novel},  motion control \cite{mellinger2011design,chen2022adaptive,chen2022aerial}, motion planning \cite{lee2016planning,alexis2016aerial,tognon2018control} and visual servoing \cite{thomas2014toward, seo2017aerial, chen2022image, ramon2020grasp} up to now. Our work focuses on the motion planning problem of aerial pick-and-place tasks, where the aerial manipulator is required to grasp and move objects in the environment (see Fig.~\ref{fig_result_aerial_pick_place}). It is noted that safety and high efficiency are important to the aerial pick-and-place task. This motivates our study to focus on an effective motion planning scheme that ensures collision-free trajectories.

Different from the motion planning of multirotors, the motion planning of an aerial manipulator is more challenging since the aerial manipulator has more degrees of freedom and is required to manipulate the objects. Different from the motion planning of a ground mobile manipulator, the motion planning of an aerial manipulator is more challenging since the aerial manipulator flies in a 3D environment rather than a 2D environment. In addition, the robotic arm and the multirotor base are dynamically coupled, which means their movements mutually affect each other. Existing approaches for motion planning for aerial manipulation can be classified into two categories based on the space in which planners calculate trajectories.

The first category is to plan the motion of the aerial manipulator in the configuration space. In early works, the RRT* method has been used to plan the path of the aerial manipulator in the configuration space without considering the dynamics \cite{lee2016planning,alexis2016aerial}. As a consequence, the resulting trajectory may not be executable for the aerial manipulator when its movement is fast. To address this issue, the dynamics of the aerial manipulator must be considered in motion planning. The existing methods that consider the dynamics in motion planning can be classified into three types. 

The first type uses a kinematics controller as a local planner in the sampling-based global planner \cite{tognon2018control}. It guarantees the feasibility of the trajectory for the real system and also enables searching for a solution directly in the reduced and more relevant task space. However, collision avoidance is not inherently embedded in the local planning, which may cause its result is not collision-free. The second type uses the differential flatness principle to ensure the dynamical feasibility \cite{ivanovic2020exploiting}. In particular, motion planning methods for a special long-reach aerial manipulator have been proposed in \cite{caballero2018first,caballero2020aerodynamic} based on this point of view. The platform in these works consists of a multirotor with a long bar extension that incorporates a lightweight dual arm in the tip. Since the dynamical feasibility constraints represented by the differential flatness are nonlinear, this type of method may be computationally expensive. The third type uses trajectory generation to ensure the dynamical feasibility \cite{kim2019sampling}. In the trajectory generation, the trajectories are represented by spline curves. The dynamical feasibility constraints are considered in the trajectory generation problem by utilizing the derivative property of the spline curves. Planning in the configuration space, however, suffers high computation costs when the dimension of the space is high \cite{lindemann2005current}. Unfortunately, aerial manipulators generally have high degrees of freedom (DoFs), which therefore motivates researchers to study other approaches to solve the motion planning problem.

The second category of approaches directly plans the trajectory of the end-effector in the Cartesian space. The motion planning of the whole aerial manipulator is often solved practically by decoupling the flying base and the manipulator \cite{garimella2021improving}. Firstly, the trajectory of the flying base approaching the manipulation position is calculated. Then, the motion of the end-effector is planned by assuming that the flying base stays in the same pose during manipulation. However, this method is conservative and inefficient in terms of energy and execution time \cite{ollero2021past}. To address this issue, the dynamic feasibility constraint must be considered in the trajectory planning of the end-effector. Therefore, a dynamically feasible task space planning method for underactuated aerial manipulators based on the differential flatness principle has been proposed in \cite{welde2021dynamically}. However, this method does not consider obstacle avoidance which is generally required in real scenarios.

\begin{figure*}[t]
	\centering
	\includegraphics[width=1\linewidth]{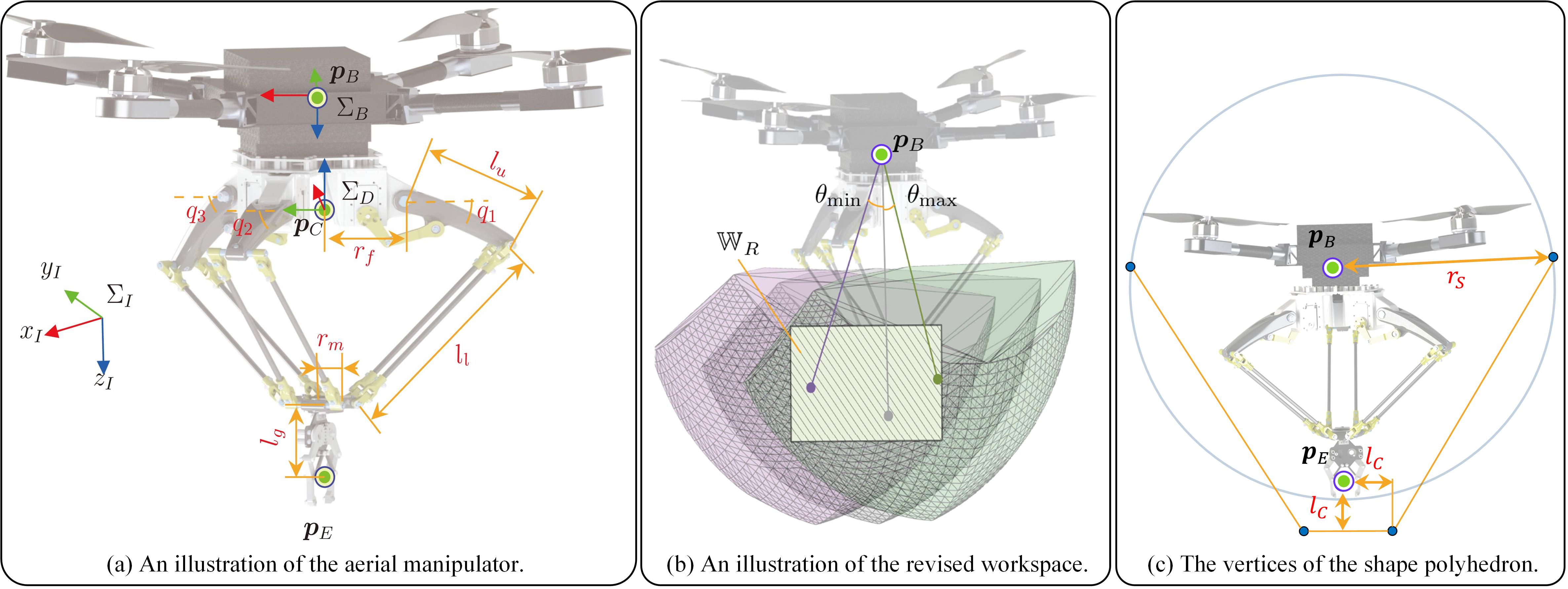}
	\caption{Coordinates, revised workspace, and shape polyhedron of the aerial manipulator.}
	\label{fig_coordinates}
\end{figure*}

The above analysis reveals the limitations of the existing motion planning approaches for aerial manipulators. Planning in the configuration space incurs high computational costs due to the high DoF of aerial manipulators, while the existing methods of planning in Cartesian space do not consider obstacle avoidance, a crucial factor in real-world scenarios. To address these limitations, this paper proposes a novel framework that integrates the motion planning of both the flying base and the manipulator in a constrained workspace. The proposed algorithm is designed for an aerial manipulator consisting of a quadcopter and a Delta arm. The novelty of our approach is outlined below:

1) We propose a novel partially decoupled motion planning method for the aerial pick-and-place task. This method calculates the dynamically feasible and collision-free trajectories of the flying base and the manipulator in Cartesian space, respectively. The resulting trajectories are coordinated for successful execution. By solving the motion planning problem in Cartesian space, the high DoF of the aerial manipulator can be handled more efficiently than planning in the configuration space with a much lower computational load. Compared with the existing methods that plan trajectories in the configuration space, this method does not suffer from the problem of increased computation in high-dimensional configuration spaces. Compared with the existing methods that plan trajectories in Cartesian space, the proposed method ensures that the trajectories are collision-free.

2) We propose novel geometric feasibility constraints to ensure the trajectories of the quadcopter and the end-effector can be successfully executed. Our proposed constraints are linearly represented by the positions of the quadcopter and the end-effector, whereas the original geometry constraints are nonlinearly represented by the configuration of the aerial manipulator. By using the constraints, our method ensures that the resulting trajectories satisfy the geometry of the aerial manipulator. This is particularly important for motion planning of the aerial manipulator in Cartesian space.

3) Collision avoidance for the Delta arm is achieved through an efficient iterative approach based on a pinhole mapping method. At each iteration, a quadratic programming (QP) problem is solved to determine the collision-free trajectory for the end-effector. A collision avoidance term, designed based on the pinhole mapping method and collision check results, is formulated into the QP problem, so that the aerial manipulator is driven away from the obstacles in the local environment. Compared to collision avoidance in the configuration space \cite{kim2019sampling,ivanovic2020exploiting}, the proposed iterative approach is faster as it is computed in Cartesian space.

The proposed algorithms are verified by three experiments on an aerial manipulator platform in the real aerial pick-and-place task. Unlike the traditional Delta arm, the Delta arm used in this paper drives the joint angles by three four-bar linkages to magnify the control forces \cite{muller1996novel}. Experiments including collision avoidance, aerial retrieval, and aerial transport are conducted to validate the novelties. 

The rest of this paper is structured as follows. The problem statement and preliminaries are given in Section~\ref{sec_2}. Kinematics and geometric feasibility constraints of the aerial manipulator are presented in Section~\ref{sec_Kinematics_Geometric}. The motion planning of the quadcopter base is proposed in Section~\ref{sec_traj_quad}. Section~\ref{sec_Delta} gives the motion planning of the Delta arm. Then, the experimental verification is given in Section~\ref{sec_experiment}. Conclusions are drawn in Section~\ref{sec_con}.

\section{Problem Statement and Preliminaries}\label{sec_2}
\subsection{Problem statement} \label{sec_problem_set}
The platform is an aerial manipulator that consists of a quadcopter and a Delta arm (see Fig.~\ref{fig_coordinates}(a)). The base of the Delta arm is attached underneath the quadcopter. The end-effector used in this paper is a gripper and it is mounted on the end of the Delta arm, whose position can be controlled by the three actuators attached to the base of the Delta arm. The orientation of the end-effector is set the same as the orientation of the quadcopter \cite{siciliano2010robotics}.

The aerial manipulator has three reference frames: the inertial frame $\varSigma_I$, the quadcopter body-fixed frame $\varSigma_B$, and the Delta arm frame $\varSigma_D$ (see Fig.~\ref{fig_coordinates}(a)). $\varSigma_I$ is an inertial frame where the $z$-axis is in the direction of the gravity vector. $\varSigma_B$ is rigidly attached to the quadcopter base. Its origin coincides with the center of gravity of the quadcopter.  $\varSigma_D$ is rigidly attached to the Delta arm base at its geometric center $\bm{p}_C$.

Let $\bm{p}_B\in\mathbb{R}^3$ and $\bm{R}_B\in SO(3)$ denote the position of the quadcopter in $\varSigma_I$ and the rotation matrix from $\varSigma_B$ to $\varSigma_I$, respectively. Let $\bm{p}_E\in\mathbb{R}^3$ denote the position of the end-effector in $\varSigma_I$. Then, the geometric relationship between $\bm{p}_E$ and $\bm{p}_B$ can be represented as
\begin{equation}
	\bm{p}_E -\bm{p}_B = \bm{R}_B \bm{p}_E^B,
	\label{eq_geometric_cons}
\end{equation}
where $ \bm{p}_E^B \in \mathbb{R}^3$ is a function of the Delta arm's actuated joint angles $q_1,q_2,q_3$.

For a pick-and-place task, denote $\bm{p}_{O}\in\mathbb{R}^3$ and $\psi_{O}\in\mathbb{R}$ as the position and the orientation of the target object in $\varSigma_I$, respectively, whereas $\psi_E\in\mathbb{R}$ denotes the orientation angle of the end-effector. Let $t_G$ denote the time that $\bm{p}_{E}$ arrives at $\bm{p}_O$ and $t_{\text{grip}}$ the closing time of the gripper. 

The goal of the motion planning for the aerial pick-and-place is to calculate the collision-free trajectories for the quadcopter and the Delta arm to move from a starting position to a feasible grasping configuration and from that grasping configuration to the end position. Given the geometric relationship, dynamical feasibility constraints, obstacles in the environment, the start position $\bm{p}_{B,\text{start}}$, and the end position $\bm{p}_{B,\text{end}}$, the resulting trajectories must be collision-free and satisfy that $\bm{p}_{E}(t)=\bm{p}_O$ and $\psi_E(t) = \psi_{O}$ when $t\in [ t_G,t_G+t_{\text{grip}}]$.

\subsection{Overview of the proposed motion planning method} \label{sec_System_Overview}

\begin{figure*}[t]
	\centering
	\includegraphics[width=1\linewidth]{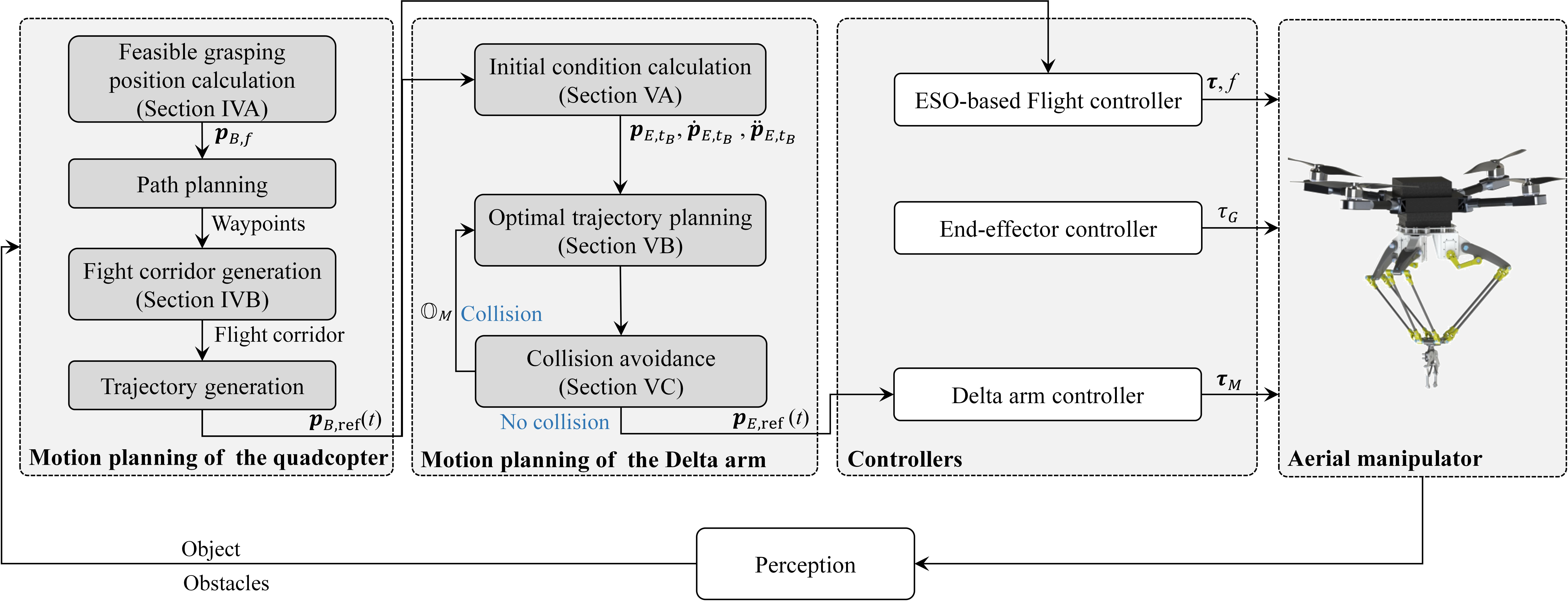}
	\caption{Structure of the proposed motion planning method for aerial pick-and-place.}
	\label{fig_system_structure}
\end{figure*}

The proposed motion planning method is partially decoupled, which calculates the trajectories of the quadcopter base $\bm{p}_{B}(t)$ and the end-effector $\bm{p}_{E}(t)$ in Cartesian space, respectively. The geometric feasibility constraints are proposed to coordinate the trajectories to ensure successful execution (see Section~\ref{subsec_Geometric} for details). The overall architecture of the motion planning and control system is shown in Fig.~\ref{fig_system_structure}.  The system is decomposed into three components.

1) The first component is the motion planning of the quadcopter base. Its inputs are the positions of the object and the obstacles. Its output is the trajectory of the quadcopter base $\bm{p}_{B,\text{ref}}(t)$. The motion planning of the quadcopter base can be further decomposed into four steps. The first step is feasible grasping position calculation. Its role is to find a suitable position for the quadcopter base to allow the aerial manipulator to grasp the object. The details of this step can be seen in Section~\ref{sec_Feasible_grasping_pos}. The second step is path planning. Its role is to find a path for the quadcopter base to move from a given starting position to a feasible grasping position and from that grasping position to a given end position. In this paper, we use the A* method to calculate the path \cite[Section~12.1.1]{nilsson2009quest}. The third step is flight corridor generation. Its role is to generate a safe flight corridor for the quadcopter base which constrains the motion of the quadcopter base to avoid collisions. The details of this step can be seen in Section~\ref{sec_flight_corridor}. The fourth step is trajectory generation. Its role is to calculate the trajectory of the quadcopter base based on the piecewise B{\'e}zier curve. We use the method proposed in \cite{gao2018online} to ensure the resulting trajectory satisfies the safety, the dynamical feasibility, and the waypoints constraints. Compared with the existing methods of aerial manipulators (e.g., \cite{kim2019sampling}), the proposed method calculates the trajectory of the quadcopter base in Cartesian space. Compared with the existing methods of the standard quadcopter (e.g., \cite{gao2018online}), the proposed method guarantees the aerial manipulator arrives at the feasible grasping configuration without collisions.  

2) The second component is the motion planning of the Delta arm. Its inputs are the position of the object $\bm{p}_O$ and $\bm{p}_{B,\text{ref}}(t)$. Its output is the trajectory of the end-effector $\bm{p}_{E,\text{ref}}(t)$. This motion planning method can be further decomposed into three steps. The first step is the initial condition calculation. Its role is to calculate the position, velocity, and acceleration of the end-effector at the beginning of the manipulation stage. The second step is the optimal trajectory planning of the end-effector based on the B{\'e}zier curve. Its role is to calculate the trajectory of the end-effector from the initial position to the object with several constraints. The trajectory planning of the end-effector is represented as a QP problem form. In particular, we propose geometric feasibility constraints of the aerial manipulator and encode this constraint into the QP problem to ensure the trajectories satisfy the geometry of the aerial manipulator. The third step is collision avoidance. It is important and its role is to ensure the trajectory of the end-effector is collision-free. In this step, the collisions between the aerial manipulator and the obstacles in a local map are detected based on the GJK method.  The second and third steps are run iteratively. If there is a collision, then the objective function of the QP problem in the second step is updated by a pinhole mapping method. Repeat the second and third steps until no collision occurs. All the corresponding sections introducing these steps are listed in Fig.~\ref{fig_system_structure}. Compared with the existing methods \cite{kim2019sampling,caballero2018motion}, the proposed method requests lower computational power since the collision avoidance of the Delta arm is achieved by an iterative approach in Cartesian space.

3) The third component is the controller of the aerial manipulator.  Its inputs are the trajectories of the quadcopter base and the end-effector. Its outputs are total force $f\in\mathbb{R}$ of the rotors, torque vector $\bm{\tau}\in\mathbb{R}^3$ of the rotors,  torque $\tau_G \in \mathbb{R}$ of the gripper, and the torque vector that each actuator should generate $\bm{\tau}_M\in \mathbb{R}^3$. The controller consists of three subcomponents. The first step is an extended state observer (ESO) -based flight controller. It was proposed in our previous work~\cite{cao2023eso} and uses ESOs to estimate dynamic coupling between the aerial manipulator and the Delta arm. Its role is to generate the force $f$ and torque vector $\bm{\tau}$ for the quadcopter base so that the trajectory of the quadcopter can be tracked. The second step is the end-effector controller. Its role is to control the gripper to grasp or release objects. The third step is the Delta arm controller. Its role is to generate the torque vector $\bm{\tau}_M$ for the Delta arm so that the trajectory of the end-effector can be tracked. The details of the Delta arm controller can be seen in our previous work~\cite{cao2023eso}. 

The steps can be classified into offboard processes and onboard processes. In Fig.~\ref{fig_system_structure}, steps in the small grey rectangle are done on an offboard computer, while processes in the white rectangle run onboard the aerial manipulator during flights.

\subsection{preliminaries to B{\'e}zier curves }

A $n$-th degree B{\'e}zier curve is defined by a set of control points and Bernstein polynomial bases. Let $\bm{c}_i\in\mathbb{R}^3 ,b_{i,n}(\tau)\in\mathbb{R}$ denote the $i$-th control point and Bernstein polynomial basis, respectively. Then, the $n$-th degree 3D B{\'e}zier curve is written as $\bm{B}(\tau)= \sum_{i=0}^{i=n}\bm{c}_i^T b_{i,n}(\tau)$, where 
\begin{equation}
	b_{i,n}(\tau)=\left(\begin{array}{c}
		n \\
		i \\
	\end{array}\right) \tau^i  (1-\tau)^{n-i},
\label{eq_basis}
\end{equation}
where $\tau\in[0,1]$, $\left(\begin{smallmatrix}
	n \\
	i \\
\end{smallmatrix}\right)$ is the binomial coefficient. According to \cite[Section~2.4]{prautzsch2002bezier}, the derivative of the B{\'e}zier curve can be obtained by Lemma~\ref{lemma_deri_bezier}. In addition, the B{\'e}zie curve B(t) is entirely confined within the convex hull defined by all these control points, which is referred to as the convex hull property (see Lemma~\ref{lemma_conv_hull}).  
\begin{lemma}[Derivative \cite{prautzsch2002bezier}]\label{lemma_deri_bezier}
	Let $\bm{B}^{(k)}(\tau)= \sum_{i=0}^{n-j}\bm{c}_{i}^{(k)}b_{i,n-j}(\tau)$ denote the $k$-th derivative of $\bm{B}(\tau)$, then the control points of $\bm{B}^{(k)}(\tau)$ can be calculated iteratively by $\bm{c}_{i}^{(k)}= (n-k + 1)(\bm{c}_{i+1}^{(k-1)}-\bm{c}_{i}^{(k-1)})$, where $i= 0, 1, \cdots, n-j$.  
\end{lemma}

\begin{lemma}[Convex hull property \cite{prautzsch2002bezier}]\label{lemma_conv_hull}
	Let $\mathbb{H}=\{ a_0\bm{c}_0 + a_1\bm{c}_1 + \cdots + a_n\bm{c}_n| a_0+a_1+\cdots+a_n =1, a_i\geq 0\}$ denote the convex hull defined by all the control points, then $\bm{B}(\tau)\in \mathbb{H} $ for all $\tau\in[0,1]$.
\end{lemma}

\section{Kinematics and Geometric Feasibility Constraints}\label{sec_Kinematics_Geometric}
This section proposes the kinematics and geometric feasibility constraints of the aerial manipulator.

\subsection{Kinematics of the aerial manipulator}\label{sec_Description}
According to \eqref{eq_geometric_cons}, the time derivative of $\bm{p}_E$ is
\begin{equation}
	\begin{split}
		\dot{\bm{p}}_E=&\dot{\bm{p}}+\dot{\bm{R}}_B\bm{p}_E^B+\bm{R}\dot{\bm{p}}_E^B\\
		=&\dot{\bm{p}} + \bm{R}_B\bm{R}_D^B\dot{\bm{p}}_E^D- [ \bm{R}_B\bm{p}_E^B]_{\times}\bm{\omega},\\
	\end{split}
	\label{eq_diffKine}
\end{equation}
where $\bm{\omega}\in\mathbb{R}^3$ is the angular velocity vector of the quadcopter expressed in $\varSigma_B$, and $[\cdot]_{\times}$ denotes the skew-symmetric matrix. 

Let $\bm{p}_C^B \in \mathbb{R}^3$ denote the position of the center of the base in $\varSigma_B$. Let $\bm{p}_E^B\in \mathbb{R}^3$ and $\bm{p}_E^D\in \mathbb{R}^3$ denote the positions of the end-effector in $\varSigma_B$ and $\varSigma_D$, respectively. The relationship between $\bm{p}_E^B$ and $\bm{p}_E^D$ is
\begin{equation}
	\bm{p}_E^B = \bm{R}_D^B\bm{p}_E^D+\bm{p}_C^B,
	\label{eq_kine_delta}
\end{equation}
where $\bm{R}_D^B\in SO(3)$ is the rotation matrix from $\varSigma_D$ to $\varSigma_B$.

The lengths for the upper and lower arms are represented by $l_U$ and $l_L$ as illustrated in Fig.~\ref{fig_coordinates}(a). Circumradius of the top base and the bottom end-effector base are, respectively, defined as $r_F$ and $r_M$. The length of the gripper is denoted as $l_g$. The relationship between the end-effector position $\bm{p}_E^D$ and the joint vector $\bm{q}=[q_1, q_2, q_3]^T\in \mathbb{R}^3$ is
\begin{equation}
	\left\| \bm{p}_E^D + \bm{l}_G-\bm{h}_i\right\|^2=l_L^2, \quad i=1,2,3,
	\label{eq_forward_kine}
\end{equation}
where $\bm{l}_G = [0,0,l_g]^T$, and
\begin{equation}
	\bm{h}_i=\left[
	\begin{array}{c}
		-(r_F-r_M+l_U \cos q_i)\cos [(i-1)\pi /3]\\
		(r_F-r_M+l_U\cos q_i)\sin [(i-1)\pi /3]\\
		l_U\sin q_i\\
	\end{array}
	\right].
\end{equation}
On the one hand, given a joint vector $\bm{q}$, the position $\bm{p}_E^D$ can be solved from \eqref{eq_forward_kine} based on the forward kinematics. On the other hand, given a position $\bm{p}_E^D$, the joint vector $\bm{q}$ can be solved from \eqref{eq_forward_kine} by the inverse kinematics. Details can be found in \cite{lopez2006delta,yang2014geometric}.

As can be seen from Fig.~\ref{fig_coordinates}(a), the joint angles of the Delta arm are driven by planar four-bar linkages. The relationship between the joint angles and the crank position angles can be calculated by the kinematics of the planar four-bar linkage \cite[Section~3.6]{shigley2004standard}.

\subsection{Geometric feasibility constraints}\label{subsec_Geometric}
Combining \eqref{eq_geometric_cons} and \eqref{eq_kine_delta}, the geometric relationship between the end-effector and the quadcopter is 
\begin{equation}
	\bm{p}_E - \bm{p}_B= \bm{R}_{B}(\bm{R}_D^B\bm{p}_E^D+\bm{p}_C^B),\quad \bm{p}_E^D\in\mathbb{W},
	\label{eq_pB_range_1}
\end{equation}
where $\mathbb{W}$ is the workspace of the Delta and can be calculated by the forward kinematics of the Delta arm. The workspace is approximated as a convex polyhedron \cite{laribi2007analysis}. Therefore, the expression of the workspace is 
\begin{equation}
	\mathbb{W}= \{ \bm{p} | \bm{A}_D\bm{p}\leq\bm{b}_{D}\}.
	\label{eq_workspace}
\end{equation}

According to \eqref{eq_pB_range_1}, the range of $\bm{p}_E - \bm{p}_B$ is determined by $\mathbb{W}$ and $\bm{R}_{B}$. We define $\bm{R}_{B} = \bm{R}_{\psi}\bm{R}_{\theta,\phi}$, where $\bm{R}_{\psi}$ is the rotation matrix determined by the yaw angle $\psi$, $\bm{R}_{\theta,\phi}$ are the rotation matrix determined by the pitch angle $\theta$ and roll angle $\phi$. The yaw angle of the quadcopter is constant, i.e., $\psi= \psi_O$, when the aerial manipulator is grasping or placing an object. Then, \eqref{eq_pB_range_1} is rewritten as 
\begin{equation}
	\bm{R}_{\psi_O}^T(\bm{p}_E - \bm{p}_B)= \bm{R}_{\theta,\phi}(\bm{R}_D^B\bm{p}_E^D+\bm{p}_C^B),\quad \bm{p}_E^D\in\mathbb{W}.
	\label{eq_pB_range_2}
\end{equation}
To make the above equation more concise, we define 
\begin{equation}
	\mathbb{W}_{\theta,\phi} = \{ \bm{R}_{\theta,\phi}(\bm{R}_D^B\bm{p}_E^D+\bm{p}_C^B)| \bm{p}_E^D\in\mathbb{W}\}.
	\label{eq_W_theta_phi_0}
\end{equation}
 Therefore, \eqref{eq_pB_range_2} is rewritten as $\bm{R}_{\psi_O}^T(\bm{p}_E - \bm{p}_B) \in\mathbb{W}_{\theta,\phi}$.

To linearize the geometric relationship \eqref{eq_pB_range_2}, we define $\mathbb{W}_R = \{ \bm{p}| \bm{w}_{\min}\leq\bm{p}\leq\bm{w}_{\max}\}$ as the revised workspace, and it satisfy that $\mathbb{W}_R \subset \mathbb{W}_{\theta,\phi}$. Since the roll and pitch angles of the quadcopter are small when the aerial manipulator is manipulating, the bounds of $\theta$ and $\phi$ can be determined with several experiments. Let $\theta_{\min}, \theta_{\max}$ denote the minimum and maximum of $\theta$.  Let $\phi_{\min}, \phi_{\max}$ denote the minimum and maximum of $\phi$. We calculate $\mathbb{W}_R$ by two steps.

The first step is calculating boundaries of $\mathbb{W}_{\theta,\phi}$. Combining \eqref{eq_workspace} and \eqref{eq_W_theta_phi_0}, the expression of $\mathbb{W}_{\theta,\phi}$ can be rewritten as 
\begin{equation}
	\mathbb{W}_{\theta,\phi} =  \{ \bm{p} | \bm{A}_D(\bm{R}_{\theta,\phi}\bm{R}_D^B )^T\bm{p}\leq\bm{b}_{D}+\bm{A}_D\bm{R}_D^{BT}\bm{p}_C^B\}. 
	\label{eq_W_theta_phi}
\end{equation}
According to \eqref{eq_W_theta_phi}, we obtain the boundaries $\mathbb{W}_{\theta = \theta_{\min},\phi = 0}$, $\mathbb{W}_{\theta = \theta_{\max},\phi = 0}$, $\mathbb{W}_{\theta = 0,\phi = \phi_{\min}}$, $\mathbb{W}_{\theta = 0,\phi = \phi_{\max}}$. 

The second step is calculating the intersection of these sets $\mathbb{W}_{I}$. According to the definition of the intersection, we have 
	\begin{equation}
		\begin{split}
				\mathbb{W}_{I} =\{ \bm{p} | \bm{A}_D(\bm{R}_{\theta_{\min},0}\bm{R}_D^B )^T\bm{p}&\leq\bm{b}_{D}+\bm{A}_D\bm{R}_D^{BT}\bm{p}_C^B,	\\
											\bm{A}_D(\bm{R}_{\theta_{\max},0}\bm{R}_D^B )^T\bm{p}&\leq\bm{b}_{D}+\bm{A}_D\bm{R}_D^{BT}\bm{p}_C^B,\\
											\bm{A}_D(\bm{R}_{0,\phi_{\min}}\bm{R}_D^B )^T\bm{p}&\leq\bm{b}_{D}+\bm{A}_D\bm{R}_D^{BT}\bm{p}_C^B,\\
											\bm{A}_D(\bm{R}_{0,\phi_{\max}}\bm{R}_D^B )^T\bm{p}&\leq\bm{b}_{D}+\bm{A}_D\bm{R}_D^{BT}\bm{p}_C^B\}.
				\label{eq_intersection}
		\end{split}
	\end{equation}
	 Since the expression of the intersection \eqref{eq_intersection} is complicated, it may be inconvenient when applied to real systems. We let the largest cuboid that can be inscribed within the intersection as $\mathbb{W}_R$. The cuboid can be calculated by the method proposed in \cite{mondal2021finding}. Then, $\bm{w}_{\min}=[{w}_{x,\min},{w}_{y,\min},{w}_{z,\min}]^T$ and $\bm{w}_{\max}=[{w}_{x,\max},{w}_{y,\max},{w}_{z,\max}]^T$ are subsequently determined by the size of the cuboid. Fig.~\ref{fig_coordinates}(b) gives an illustration for calculating the revised workspace in the pitch direction. Then, the geometric feasibility constraints are 
\begin{equation}
		\bm{w}_{\min}\leq\bm{R}_{\psi_O}^T(\bm{p}_{E}-\bm{p}_{B})\leq\bm{w}_{\max},
		\label{eq_geom_const_simp}
\end{equation} 
where
 \begin{equation}
	\bm{R}_{\psi_{O}}=\left[
	\begin{array}{ccc}
		\cos\psi_{O} & -\sin\psi_{O} & 0\\
		\sin\psi_{O} &  \cos\psi_{O} & 0\\
		0 & 0 & 1\\
	\end{array}
	\right].
\end{equation}

\section{Motion Planning for the Quadcopter base}\label{sec_traj_quad}
This section presents a method to generate the trajectory of the quadcopter for the aerial pick-and-place task. This method consists of four steps: feasible grasping position, path planning, flight corridor generation, and B{\'e}zier curve-based trajectory generation. The purposes and relationships of these steps are given in Section~\ref{sec_System_Overview}. In the algorithm, the path planning can be achieved by the existing method. In our work, we use the A* method to obtain the path in the 3D grid map which is used to represent the environment of the task. The B{\'e}zier curve-based trajectory generation is achieved by an existing method proposed in \cite{gao2018online}. It bounds positions and higher order dynamics of the trajectory entirely within safe regions by using Bernstein polynomial basis and formulating the trajectory generation problem as typical convex programs.

Compared to the existing methods for standard quadcopters \cite{gao2018online}, the proposed method for the quadcopter base of the aerial manipulator has two novelties. First, the feasible grasping position is calculated to ensure the aerial manipulator can manipulate the object. Second, the volume of the aerial manipulator changes with the movement of the Delta arm. To address this issue, the aerial pick-and-place task is divided into two stages: moving and manipulation stages. The flight corridors in the two stages are obtained, respectively. The details are shown as follows.

\subsection{Feasible grasping position}\label{sec_Feasible_grasping_pos}
  To grasp the object, the position of the end-effector must arrive at $\bm{p}_{O}$ with an orientation angle of  $\psi_{O}$. The feasible grasping position of the quadcopter is constrained by the geometric shape of the aerial manipulator. Let $\bm{p}_{B,f}\in\mathbb{R}^3$ denote the feasible grasping position. Let  $\bm{R}_{B,f}\in SO(3)$ denote the desired rotation matrix of the quadcopter base at the feasible grasping position. According to \eqref{eq_geometric_cons}, the feasible grasping position of the quadcopter is  
 \begin{equation}
 	\bm{p}_{B,f}=\bm{p}_O - \bm{R}_{B,f}\bm{p}_E^B,
 	\label{eq_pB_d_1}
 \end{equation}
  According to \eqref{eq_pB_d_1}, one can conclude that $\bm{R}_{B,f}$ and $\bm{p}_E^B$ need to be determine before calculating $\bm{p}_{B,f}$. 
 
   In the manipulation stage, the yaw angle of the quadcopter is set as $\psi_{O}$ to satisfy the grasp angle constraint of the end-effector. We assume that the roll and pitch angles of the quadcopter are small when the quadcopter base is around $\bm{p}_{B,f}$. This assumption is reasonable since the motion of the quadcopter is conservative. According to the assumption, we have $\bm{R}_{B,f} =\bm{R}_{\psi_{O}}$. To ensure the manipulability of the Delta arm, we let the end-effector stay at the center of $\mathbb{W}_R$ when the aerial manipulator picks up the object. Then, \eqref{eq_pB_d_1} is rewritten as    
 \begin{equation}
 	\bm{p}_{B,f} = \bm{p}_{O} - 0.5\bm{R}_{\psi_O}(\bm{w}_{\min}+\bm{w}_{\max}).
 	\label{eq_pid1}
 \end{equation}
 For an aerial pick-and-place task, we already have the start position $\bm{p}_{B,\text{start}}$, feasible grasping position $\bm{p}_{B,f}$, and the end position $\bm{p}_{B,\text{start}}$. Then, the path of the quadcopter can be obtained by the A* method. 

\subsection{Flight corridor generation}\label{sec_flight_corridor}
  The flight corridor is a collection of convex overlapping polyhedra that models free space and provides a connected corridor containing the resulting path. A convex decomposition method proposed in \cite{liu2017planning} is adopted to generate the flight corridor by inflating the resulting path. However, this method was originally designed for a traditional quadcopter with a fixed volume, while the volume of the aerial manipulator changes with the movement of the Delta arm. Therefore, the method cannot be directly used for the aerial manipulator. To address this issue, we calculate the flight corridors in the moving and the manipulation stages, respectively. In the moving stage, the position of the end-effector is set to stay at the top point $\bm{p}_{\text{top}}^B\in\mathbb{R}^3$ of the Delta arm's workspace $\mathbb{W}_R$. From the definition of $\mathbb{W}_R$, we have  
	\begin{equation}
	 	\bm{p}_{\text{top}}^B = \left[ 
	 	\begin{Array}{c}
	 		0.5(w_{x,\min}+w_{x,\max})\\
	 		 0.5(w_{y,\min}+w_{y,\max})\\
	 		 w_{z,\min}
	 	\end{Array}
	 	\right].
		\label{eq_p_top_B}
	\end{equation}
 The shape of the aerial manipulator now can be approximated as a sphere with a radius $r_S$. Then, we can use the convex decomposition method to generate the flight corridor in the moving stage.

\begin{figure*}[t]
	\centering
	\includegraphics[width=1\linewidth]{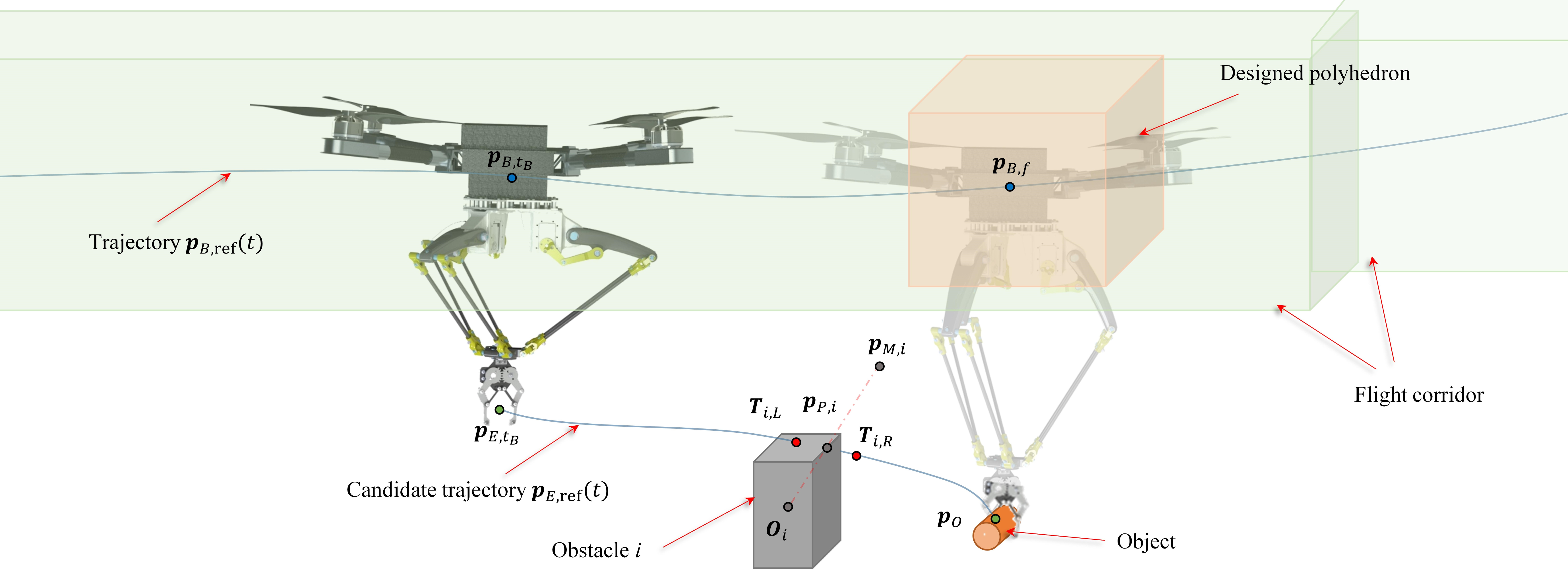}
	\caption{An illustration for the motion planning of the aerial pick-and-place.}
	\label{fig_two_polyhedron}
\end{figure*}

In the manipulation stage, we use a designed polyhedron as the flight corridor to ensure the object is reachable for the aerial manipulator (see Fig.~\ref{fig_two_polyhedron}). 
The designed polyhedron is designed based on the geometric feasibility constraints \eqref{eq_geom_const_simp} and it is represented as
\begin{equation}
	\bm{w}_{\min}\leq \bm{R}_{\psi_{O}}(\bm{p}-\bm{p}_{B,d}) \leq \bm{w}_{\max}.
\end{equation}
The duration time in this polyhedron is determined by the mechanical behavior of the gripper. We set the duration time as the closing time of the gripper $t_{\text{grip}}$.

\section{Motion Planning for The Delta Arm}\label{sec_Delta}
In this section, we calculate the collision-free trajectory of the Delta arm in the Cartesian coordinate. The proposed method for the Delta arm utilizes the resulting trajectory of the quadcopter. In the moving stage, the Delta arm stays at an initial state and its end-effector stays at a fixed position $\bm{p}_{\text{top}}^B$ relative to the quadcopter base. The position $\bm{p}_{\text{top}}^B$ can be calculated by \eqref{eq_p_top_B}. Therefore, the Delta arm does not require additional motion planning calculations in the moving stage.

 Let $t_{B}$ denote the time at which the quadcopter base enters the designed polyhedron. Let $t_{B}$ denote the beginning time of the manipulation stage. The time  $t_{B}$ can be determined by $t_B=t_G-2v_{E,\max}/a_{E,\max}$, where $v_{E,\max}$ and $a_{E,\max}$ are maximum velocity and acceleration of the end-effector, respectively. The procedure of the manipulation stage is given here. From $t_{B}$ to $t_G$, the end-effector moves to the object. Then, the aerial manipulator keeps the position of the end-effector for the duration time $t_{\text{grip}}$ to pick up or place the object. After picking up or placing the object, the Delta arm returns to its initial state. According to the procedure, one can find that the trajectory of the end-effector from $t_{B}$ to $t_G$ needs to be calculated. The details to calculate the trajectory from $t_{B}$ to $t_G$ are shown as follows.

\subsection{Initial condition}\label{sec_initial_condition}
The initial condition for the end-effector consists of the initial position $\bm{p}_{E,t_B}$, the initial velocity $\dot{\bm{p}}_{E,t_B}$, and the initial acceleration $\ddot{\bm{p}}_{E,t_B}$. They are calculated as follows.

\subsubsection{Initial position}
According to \eqref{eq_geometric_cons}, the initial position $\bm{p}_{E,t_B}$ is calculated by 
\begin{equation}
	\bm{p}_{E,t_B}=\bm{p}_{B,t_B} + \bm{R}_{B,t_B}\bm{p}_{\text{top}}^B, 
	\label{eq_p_e_tb}
\end{equation}
where $\bm{p}_{B,t_B}, \bm{R}_{B,t_B}= [\bm{r}_{1,t_B},\bm{r}_{2,t_B},\bm{r}_{3,t_B}]$ denote $\bm{p}_{B}, \bm{R}_{B}$ at the time $t_B$, respectively. According to \eqref{eq_p_e_tb}, we calculate $\bm{p}_{B,t_B}$ and $\bm{R}_{B,t_B}$ to obtain $\bm{p}_{E,t_B}$. $\bm{p}_{B,t_B}$ can be directly obtained by the trajectory of the quadcopter.  

The matrix $\bm{R}_{B,t_B}$ is calculated based on the differential flatness of the quadcopter. At the time $t_B$, the yaw angle of the quadcopter base is $\psi_{O}$ to ensure the orientation angle of the end-effector equals $\psi_{O}$. The unit orientation vector in the ground plane is $\bm{r}_g = [\cos\psi_O, \sin \psi_O,0]^T$. According to \cite{mellinger2011minimum}, we have 
\begin{equation}
	\bm{r}_{3,t_B}= \frac{\ddot{\bm{p}}_{t_{k}}+g\bm{e_3}}{\| \ddot{\bm{p}}_{t_B}+g\bm{e_3}\| }, 
	\label{eq_r3}
\end{equation}
and vectors $\bm{r}_{1,t_B}$ and $\bm{r}_{2,t_B}$ can be determined by
\begin{equation}
	\bm{r}_{2,t_B} = \frac{\bm{r}_{3,t_B}\times \bm{r}_g}{\| \bm{r}_{3,t_B}\times \bm{r}_g\|}, \bm{r}_{1,t_B} = \bm{r}_{2,t_B} \times \bm{r}_{3,t_B}. 
	\label{eq_r1_r2}
\end{equation} 

\subsubsection{Initial velocity and acceleration}
Let $\delta_{I}$ denote a small time step. We can calculate $\bm{p}_{E,t_B-\delta_{I}}$ and $\bm{p}_{E,t_B+\delta_{I}}$ according to the above method. Then, the derivatives are approximated by the differences. The initial velocity and acceleration are 
\begin{equation}
	\begin{split}
		\dot{\bm{p}}_{E,t_B}  &= (\bm{p}_{E,t_B} - \bm{p}_{E,t_B-\delta_{I}})/\delta_{I},\\
		\ddot{\bm{p}}_{E,t_B} &= (\bm{p}_{E,t_B+\delta_{I}} - 2\bm{p}_{E,t_B} +\bm{p}_{E,t_B-\delta_{I}})/\delta_{I}^2.
	\end{split}
\end{equation}

\subsection{Optimal trajectory planning}\label{sec_Optimal_trajectory_planning}
 The trajectory of the end-effector is calculated by an iterative approach. The trajectory planning of the end-effector is formulated as a QP problem. At each iteration, the objective function of the QP problem is updated and the QP problem is solved to calculate the collision-free trajectory of the end-effector. Let $\bm{p}_{E,\text{ref}}(t), t\in[t_B, t_G]$ denote the trajectory. A $n_E$-th order B{\'e}zier curve is adopted to represent the trajectory and it is 
  \begin{equation}
  	\bm{p}_{E,\text{ref}}(\tau_M) =\sum_{i=0}^{n_E}\bm{c}_{E,i}b_{i,n_E}(\tau_M),
  	\label{eq_define_p_e_ref}
 \end{equation}
where $\bm{c}_{E,i}=[{c}_{E,x,i},{c}_{E,y,i},{c}_{E,z,i}]^T\in\mathbb{R}^3$ and ${b}_{i,n_E}(\tau_M)$ are the $i$-th control point and Bernstein polynomial basis of the B{\'e}zier curve, respectively, and $\tau_M = (t-t_B)/(t_G-t_B)$. 
We denote the parameter vector of the trajectory as $\underline{\bm{c}}_E = [{c}_{E,x,0},\ldots, {c}_{E,x,n_E},{c}_{E,y,0}, \ldots,{c}_{E,y,n_E},{c}_{E,z,0},\ldots,{c}_{E,z,n_E}]^T$. The QP problem is then formulated as   
\begin{equation}
	\begin{split}
		\min \quad &J = \underline{\bm{c}}_E^T\bm{Q}_{O,E}\underline{\bm{c}}_E + \bm{q}_{O,E}^T\underline{\bm{c}}_E\\
		\text{s.t.}\quad & \bm{A}_{E,eq}\underline{\bm{c}}_E = \bm{b}_{E,eq},\\
		&\bm{A}_{E,ie}\underline{\bm{c}}_E\leq \bm{b}_{E,ie}, \\
	\end{split}
	\label{eq_qp_end}
\end{equation} 
where $\bm{Q}_{O,E} \in \mathbb{R}^{3(n_E+1)\times 3(n_E+1)}$ is the Hessian matrix of the objective function and it is semidefinite, $\bm{q}_{O,E}\in\mathbb{R}^{3(n_E+1)}$ is a vector, $\bm{A}_{E,eq}\in \mathbb{R}^{18\times 3(n_E+1)}$ and $\bm{A}_{E,ie}\in\mathbb{R}^{(9n_{E}+4)\times 3(n_E+1)}$ are constraint matrices, $\bm{b}_{eq} \in \mathbb{R}^{18}$ and $\bm{b}_{ie}\in\mathbb{R}^{9n_{E}+4}$ are constraint vectors. The linear equality constraint ($\bm{A}_{E,eq} = \bm{b}_{E,eq}$) is endpoint constraints. The linear inequality constraint ($\bm{A}_{ie} = \bm{b}_{ie}$) consists of dynamical feasibility, geometric feasibility, and grasp constraints. These constraints are adopted to ensure the solution of the problem~\eqref{eq_qp_end} is collision-free and can be executed successfully. Definitions and roles of the objective and the constraints are as follows.

\subsubsection{Objective function}
The objective function is denoted as $J= J_J +  J_O$, where $J_J$ is the cost function to minimize the jerk along the trajectory, and $J_O$ is a penalty function for the collision. The details of the two terms are  
\begin{equation}
	J_J=\sum_{i=1}^{m}\int_{T_{i-1}}^{T_{i-1}}(j_x^{2}(t) +j_y^{2}(t) + j_z^{2}(t))dt,
	\label{eq_jj}
\end{equation}
\begin{equation}
	J_O = \sum_{k=1}^{n_O}\lambda_k\sum_{i=0}^{n_E}(c_{i,x}-x_{M,k})^2+(c_{i,y}-y_{M,k})^2+(c_{i,z}-z_{M,k})^2,
	\label{eq_jo}
\end{equation}
where $j_x, j_y, j_z$ denote the jerks of the trajectory in the corresponding three dimensions, respectively, $\lambda_k$ is a changing weighting factor, $x_{M,k},y_{M,k},z_{M,k}$ are corresponding elements of the obstacle mirror position $\bm{p}_{M,k}$. We define the obstacle mirror set as $\mathbb{O}_M=\{\bm{p}_{M,1}, \bm{p}_{M,2},\ldots, \bm{p}_{M,n_O}\}$, where $n_O$ is the number of the obstacles that collide with the aerial manipulator during the whole iteration process. The obstacle mirror position $\bm{p}_{M,k}$ can be obtained through pinhole mapping of the corresponding obstacle position. As the iterations progress, the algorithm can guide the trajectory of the end-effector towards the obstacle mirror positions while ensuring that it is collision-free for the obstacles. See Section~\ref{sec_collision_check} for details of calculating the changing weighting factors $\lambda_1, \lambda_2,\ldots, \lambda_{n_O}$ and the obstacle mirror set $\mathbb{O}_M$. By using the Lemma~\ref{lemma_deri_bezier} into \eqref{eq_jj}, we can obtain 
$J = \underline{\bm{c}}_E^T\bm{Q}_{o,E}\underline{\bm{c}}_E + \bm{q}_{o,E}^T\underline{\bm{c}}_E$. We leave the details of the $\bm{Q}_{o,E}$ and $\bm{q}_{o,E}$ for brevity.

\subsubsection{Constraints}
The constraints for the trajectory planning problem of the end-effector consist of endpoint, dynamical feasibility, geometric feasibility, and grasp constraints. The details of the constraints are given as follows:

The endpoint constraints are introduced to ensure the trajectory of the end-effector starts at $\bm{p}_{E,t_B}$ and ends at $\bm{p}_{O}$ with desired velocities and accelerations. The endpoint constraints are given as 
\begin{equation}
	\begin{split}
		c_{0,\mu,E} &= \mu_{E,t_B},s_E^{-1} c_{0,\mu,E}^{(1)}= \dot{\mu}_{E,t_B}, s_E^{-2}c_{0,\mu,E}^{(2)}= \ddot{\mu}_{E,t_B},\\
		c_{n_E,\mu,E} &= \mu_{O}, s_E^{-1} c_{n_E,\mu,E}^{(1)} = \dot{\mu}_{E,t_G}, s_E^{-2}c_{n_E,\mu,E}^{(2)}  = \ddot{\mu}_{E,t_G},\\
	\end{split}
\end{equation}
where $c_{i,\mu,E}^{(k)}$ denotes the $i$-th control point of ${d^kf_{\mu}(\tau)}/{d\tau^k}$ and can be calculated by Lemma~\ref{lemma_deri_bezier}, $\mu \in [x, y, z]$, $s_E = t_G-t_B$, $\bm{p}_{E,t_B}, \dot{\bm{p}}_{E,t_B}, \ddot{\bm{p}}_{E,t_B}$ can be obtained from the Section~\ref{sec_initial_condition} and  $\dot{\bm{p}}_{E,t_B} = 0, \ddot{\bm{p}}_{E,t_B} = 0$

The dynamical feasibility constraints consist of velocity and acceleration constraints to ensure the generated trajectory is dynamically feasible. The dynamical feasibility constraints are 
\begin{equation}
	\begin{split}
		\dot{\mu}_{\min} \leq & s_E^{-1}c_{i,\mu,E}^{(1)} \leq \dot{\mu}_{\max}, i=0,1,\ldots, n_E-1, \\
		\ddot{\mu}_{\min} \leq & s_E^{-2}c_{i,\mu,E}^{(2)} \leq \ddot{\mu}_{\max}, i=0,1,\ldots, n_E-2, \\
	\end{split}
\end{equation}
where $\mu \in [x, y, z]$, the subscript ${\min}$ denotes the lower bound of the corresponding variable, and the subscript ${\max}$ denotes the upper bound of the corresponding variable. 

The geometric feasibility constraints are introduced to ensure the trajectories of the quadcopter and the end-effector are geometrically feasible for the Delta arm. According to \eqref{eq_geom_const_simp}, it can be describe as  
\begin{equation}
	\bm{R}_{\psi_{O}}\bm{w}_{\min}\leq \bm{p}_{E,\text{ref}}(t)-\bm{p}_{B,\text{ref}}(t)=\bm{R}_{\psi_{O}}\bm{w}_{\max}.
	\label{eq_geom_cons_1}
\end{equation}
As stated above, the trajectory of the end-effector is represented by a $n_E$-th order B{\'e}zier curve. The trajectory of the quadcopter is part of a $n_B$-th order B{\'e}zier curve. 

To reveal the geometric feasibility constraints on the parameters, we use a $n_E$-th order B{\'e}zier curve to fit the trajectory of the quadcopter from $t_B$ to $t_G$. Then, the geometric feasibility constraints on the parameters can be formulated as linear algebraic equations. Let $\bm{p}_{B,0},\bm{p}_{B,1},\ldots,\bm{p}_{B,n_E}$ denote $n_E+1$ points of the trajectory $\bm{p}_{B,\text{ref}}(t), t\in[t_B, t_G]$. These points divide the trajectory into $n_E$ segments. The time interval between each two adjacent points is the same. The $n_E$-th order B{\'e}zier curve is denoted as  $\bm{h}(t) = \sum_{i=0}^{n_E}\bm{c}_{B,i}b_{i,n_E}(\tau_M)$, where $\bm{c}_{B,i}=[{c}_{B,x,i},{c}_{B,y,i},{c}_{B,z,i}]^T$ is the $i$-th control point of $\bm{h}(t)$. The control points can be obtained by fitting $\bm{h}(t)$ to the points $\bm{p}_{B,0},\bm{p}_{B,1},\ldots,\bm{p}_{B,n_E}$. Then, the geometric feasibility constraints can be rewritten as 
\begin{equation}
	\bm{R}_{\psi_{O}}\bm{w}_{\min} \leq \sum_{i=0}^{n_E}(\bm{c}_{E,i} - \bm{c}_{B,i})b_{i,n_E}(\tau_M)\leq \bm{R}_{\psi_{O}}\bm{w}_{\max},\\
	\label{eq_geom_cons_2}
\end{equation}
According to the convex hull property (see Lemma~\ref{lemma_conv_hull}), the geometric feasibility constraints on the parameters is 
\begin{equation}
	\begin{split}
		w_{r,\mu,\min} + c_{B,\mu, i} \leq & c_{E,\mu, i}\leq w_{r,\mu,\max} + c_{B,\mu, i},\\ &i=0,1,\ldots, n_E,
	\end{split}
	\label{eq_geom_cons_3}
\end{equation}
where $w_{r,\mu,\min}$ is the element of $\bm{R}_{\psi_{O}}\bm{w}_{\min}$, $w_{r,\mu,\max}$ is the element of $\bm{R}_{\psi_{O}}\bm{w}_{\max}$, and $\mu\in\{x, y, z\}$.

The grasp constraints are introduced to ensure the gripper does not collide with the object. To avoid such collision, we let the end of the trajectory in a cone. Let $t_C$ denote the time to enter the cone. Let $\bm{p}_{E,\text{ref}} (t_C) = [x_{E,t_C}, y_{E,t_C}, z_{E,t_C}]^T$ denote the position of the end-effector at the time $t_c$. Then, we have 
\begin{equation}
	\begin{split}
		  -\tan \gamma \leq \frac{x_{E,t_C} - x_O}{z_{E,t_C} - z_O} &\leq \tan \gamma,\\
		 -\tan \gamma \leq  \frac{y_{E,t_C} - y_O}{z_{E,t_C} - z_O}  &\leq \tan \gamma,\\
	\end{split}
	\label{eq_grasp_constraint_1}
\end{equation} 
where $\gamma$ is the angle of the cone. By substituting the defined \eqref{eq_define_p_e_ref} into \eqref{eq_grasp_constraint_1}, the grasp constraints \eqref{eq_grasp_constraint_1} can be rewritten as a linear form
\begin{equation}
	\begin{split}
		\sum_{i=0}^{n_E}({c}_{E,x,i}+{c}_{E,z,i}\tan \gamma)b_{i,n_E}(\tau_C)  \leq &  x_O + z_O\tan \gamma,\\
		\sum_{i=0}^{n_E}({c}_{E,x,i}-{c}_{E,z,i}\tan \gamma)b_{i,n_E}(\tau_C)  \geq &  x_O - z_O\tan \gamma,\\
		\sum_{i=0}^{n_E}({c}_{E,y,i}+{c}_{E,z,i}\tan \gamma)b_{i,n_E}(\tau_C) \leq  &  y_O + z_O\tan \gamma,\\
		\sum_{i=0}^{n_E}({c}_{E,y,i}-{c}_{E,z,i}\tan \gamma)b_{i,n_E}(\tau_C) \geq  &  y_O - z_O\tan \gamma,\\
	\end{split}
\label{eq_grasp_constraint}
\end{equation}
where $\tau_C = (t_C-t_B)/(t_G-t_B)$.
 
 \begin{figure*}[t]
 	\centering
 	\includegraphics[width=1\linewidth]{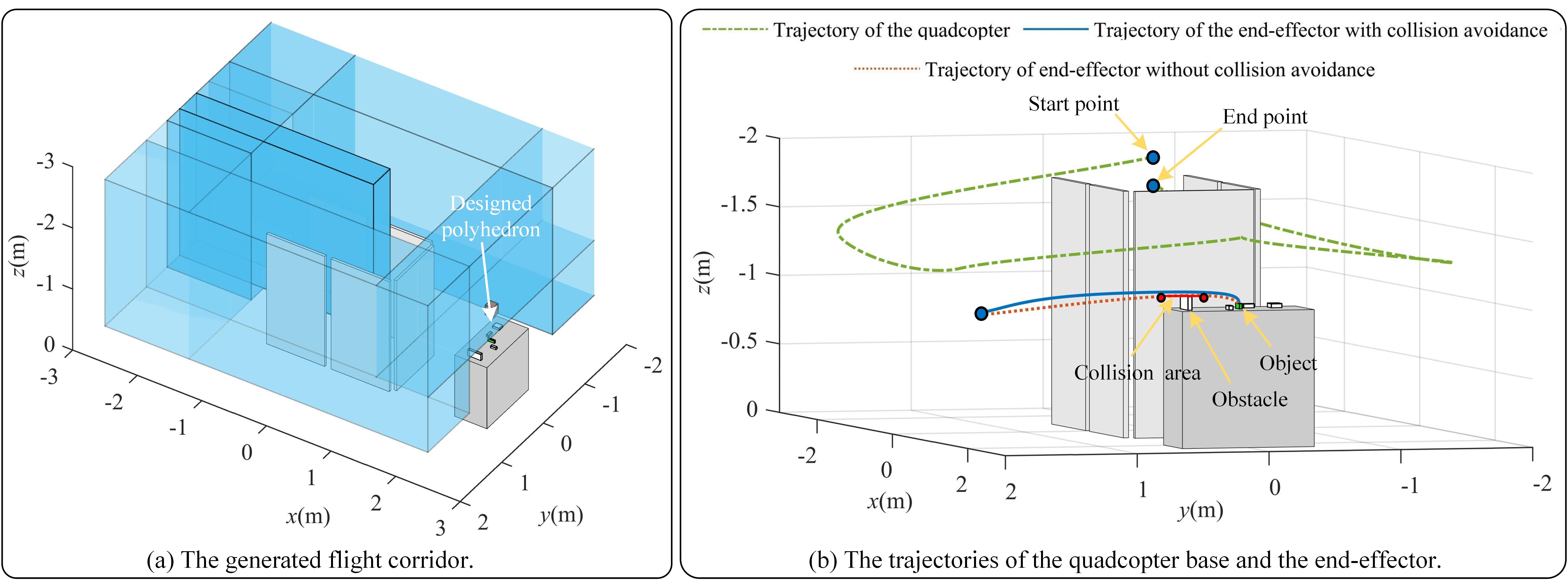}
 	\caption{Results of the collision avoidance experiment.}
 	\label{fig_sim_collision_avoidance}
 \end{figure*}
 
 \begin{figure*}[t]
 	\centering
 	\includegraphics[width=1\linewidth]{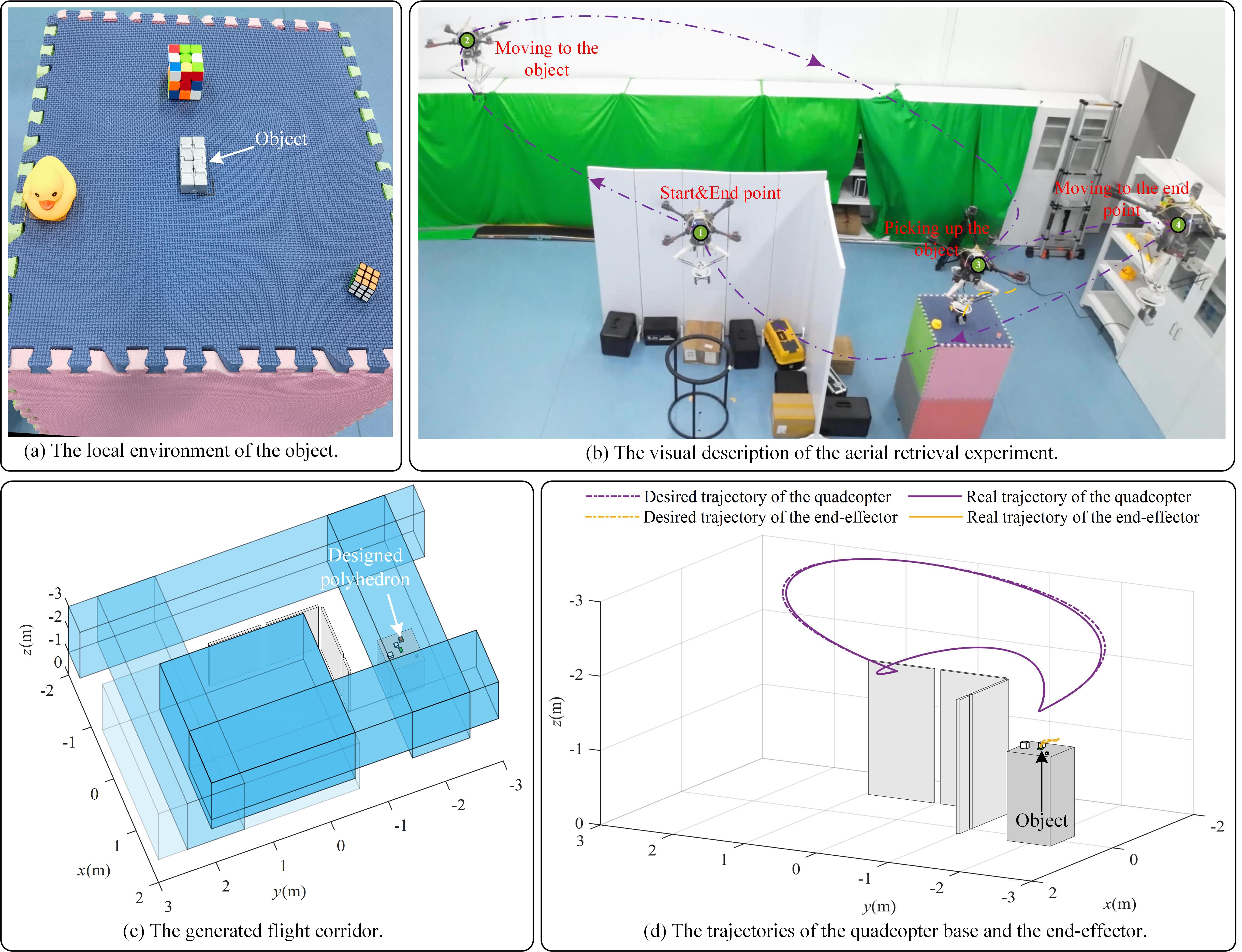}
 	\caption{Results of the aerial retrieval experiment.}
 	\label{fig_experiment_aerial_retrieval}
 \end{figure*}
 
 \begin{figure*}[t]
 	\centering
 	\includegraphics[width=1\linewidth]{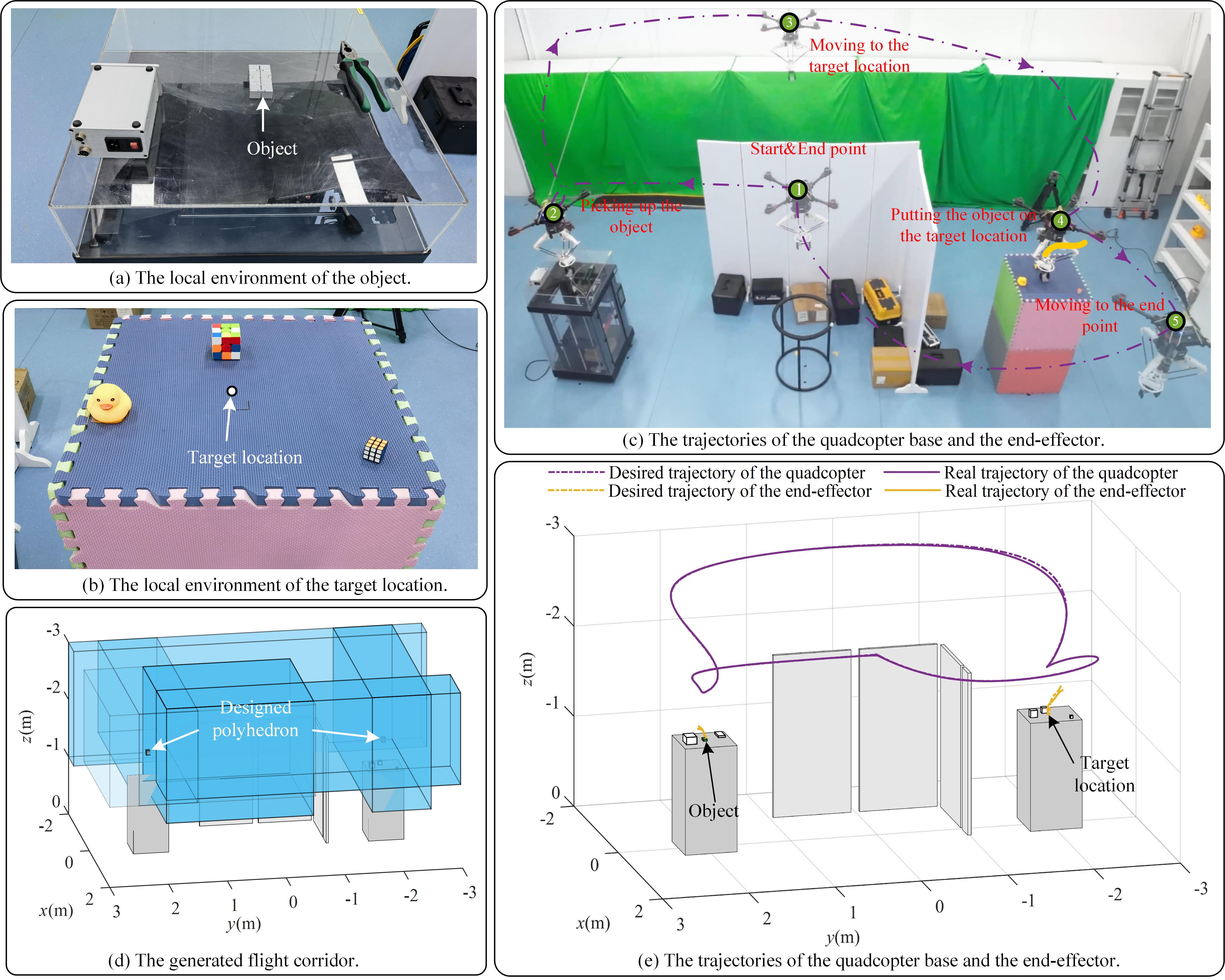}
 	\caption{Results of the aerial transport experiment.}
 	\label{fig_experiment_aerial_pick_place}
 \end{figure*}

\subsection{Collision avoidance}\label{sec_collision_check}

The subsection proposes a method to detect collisions and calculate the changing weighting factors $\lambda_1, \lambda_2,\ldots, \lambda_{n_O}$ and the obstacle mirror set $\mathbb{O}_M$ in \eqref{eq_jo}. Before the iteration process, $\mathbb{O}_M$ is set as an empty set, i.e., $\mathbb{O}_M= \emptyset$, and $n_O$ is set as zero. At each iteration, the collision is detected using the solution of the QP problem \eqref{eq_qp_end}. If the solution is collision-free, the iteration process is terminated and the collision-free solution is outputted as the trajectory of the end-effector. If there are collisions between the aerial manipulator and the obstacles in the environment with the solution, we calculate the changing weighting factors and $\mathbb{O}_M$ for the next iteration calculation.

The collision detection method is proposed to detect if the solution of the QP problem \eqref{eq_qp_end} is collision-free. The method considers the collision of the Delta arm and the end-effector. The trajectory of the quadcopter is collision-free, which is ensured by the flight corridor. Therefore, we do not consider the collision of the quadcopter. The proposed collision detection method consists of three steps.

We first use a shape polyhedron to represent the Delta arm and the end-effector in collision detection.  The vertices of the shape polyhedron are  $\bm{p}_{U,i},\bm{p}_{L,i}, i=1,2,3$ (see the blue points in Fig.\ref{fig_coordinates}(c)). Then, we have 
\begin{equation}
	\bm{p}_{U,i} = \bm{p}_{B} + \bm{R}_{\psi}\bm{R}_D^B\tilde{\bm{p}}_{U,i}, i=1,2,3,
\end{equation}
where 
\begin{equation}
	\tilde{\bm{p}}_{U,i} = \left[ 
	\begin{Array}{c}
		r_S\cos(\frac{1+2i}{3}\pi)\\
		r_S\sin(\frac{1+2i}{3}\pi)\\
		0\\
	\end{Array}
	\right] 
\end{equation}
In addition, the lower vertices can be calculated by 
\begin{equation}
	\bm{p}_{L,i} = \bm{P}_E + \bm{R}_{\psi}\bm{R}_D^B\tilde{\bm{p}}_{L,i}, i=1,2,3,
\end{equation}
where 
\begin{equation}
	\tilde{\bm{p}}_{L,i} = \left[ 
	\begin{Array}{c}
		l_C\cos(\frac{1+2i}{6}\pi)\\
		l_C\sin(\frac{1+2i}{6}\pi)\\
		l_C\\
	\end{Array}
	\right] 
\end{equation}
where $l_C$ is a constant parameter for safety and is determined by the size of the gripper when the gripper is open.

Second, we introduce a local map to reduce the computational cost of the collision detection. This is because detecting collisions in the entire environment can be computationally expensive, especially if the environment is large or if there are many obstacles. We decrease the number of obstacles to be calculated by adding a box around the end-effector and the object and thus only detecting collisions inside it. The size of the box is determined by $\bm{p}_{B,t_B}$ and $\bm{p}_{O}$. We let $\mathbb{M}_{\text{local}}=\{\bm{p}\in\mathbb{R}^3|\bm{l}_{\min}\leq \bm{p} \leq \bm{l}_{\max}\}$ denote the box. In addition, $\bm{l}_{\min}$ and $\bm{l}_{\max}$ can be calculated by
\begin{equation}
	\begin{split}
		l_{\mu,\min} = \min\{ \mu_{B,t_B}, \mu_O\}-l_s,\\
		l_{\mu,\max} = \max\{ \mu_{B,t_B}, \mu_O\}+l_s,\\
	\end{split}
	 \label{eq_box}
\end{equation}
where $\mu\in\{x,y,z\}$,  $l_{\mu,\min}, l_{\mu,\max}, \mu_{B,t_B}, \mu_O$ are corresponding element of $\bm{l}_{\min}, \bm{l}_{\min}, \bm{p}_{B,t_B}, \bm{p}_{O}$, respectively, $l_s$ is a constant parameter for safety.

Third, we use the GJK method proposed in \cite{ong2001fast} to detect collisions. If the QP problem solution reveals a collision between the aerial manipulator and obstacle $i$, then the two endpoints $\bm{T}_{i,L}$ and $\bm{T}_{i,R}$ of the solution within the collision area with respect to the obstacle can be determined (see Fig.~\ref{fig_two_polyhedron}). Let $\bm{O}_i$ denote the center of the obstacle $i$. Then, we calculate the obstacle mirror position $\bm{p}_{M,i}$ by a pinhole mapping method. The position of the pinhole is $\bm{p}_{P,i}=0.5(\bm{T}_{i,L}+\bm{T}_{i,R})$. And, we have 
\begin{equation}
	\bm{p}_{M,i} = 2\bm{p}_{P,i}-\bm{O}_i.
\end{equation}
The values of the weighting factors are updated by $\lambda_i = \lambda_i + \alpha \Delta\lambda_i$, where $i=1, 2, \ldots, n_O$, and $\alpha>0$ is a constant gain. The parameter $\Delta\lambda_i$ is the step size used for updating the $i$-th weighting factor and is a critical factor that affects the computation time of the method. The expression for $\Delta\lambda_i$ is given as 
\begin{equation}
	\Delta\lambda_i = \|\bm{T}_{i,L}-\bm{T}_{i,R} \|.
\end{equation} 

\section{Experimental Verification}\label{sec_experiment}
This section presents experimental results to verify the effectiveness of the proposed motion planning algorithms. The experimental video is available at https://youtu.be/q7O9v7l2Oho.

First of all, we describe the experimental setup. The aerial manipulator platform used in the experiments consists of a quadcopter and a Delta arm. The wheelbase of the quadcopter is $0.65$~m. The mass of the quadcopter (including a battery) is $3.60$~kg. The Delta arm consists of a mounting base ($0.56$~kg), a movable robotic arm ($0.44$~kg), and a gripper ($0.32$~kg). A flight controller proposed in our previous work~\cite{cao2023eso} runs on a Pixhawk 4 autopilot. This controller uses extended state observers (ESOs) to estimate dynamic coupling between the aerial manipulator and the Delta arm. The proposed motion planning method runs on an onboard Intel NUC i7 computer with ROS (an open-source robotics middleware suite). The experiments are conducted in a Vicon system, which provides accurate position measurements of the quadcopter base and the end-effector. The measurement data of the Vicon system is sent to a ground control station through an ethernet switch. Then, the ground control station sends the measurement data to the aerial manipulator with a frequency of 100 Hz through a 5 GHz~ wireless router.

The perception of the aerial manipulator is not surveyed in this paper. We assume that the obstacles in the environment are already known. In particular, the locations of the obstacles and the object can be obtained by the Vicon system. Then, the environment can be previously built as a grid map which consists of a set of cubes. The size of each cube is set as 0.1 m. This map is used for the path planning of the quadcopter base. The description of the controllers is provided in Section~\ref{sec_System_Overview}.

In all the examples, we use the same set of parameters of the motion planner: $\alpha = 3.0$, $r_S=0.50$~m, $l_C=0.06$~m, $l_S = 0.20$~m. The velocity and the acceleration constraints for the quadcopter base are set as 0.5 m/s and 1.0 m/s$^2$, respectively. The velocity and the acceleration constraints for the end-effector are set as 0.5 m/s and 2.0 m/s$^2$, respectively. The bounds of the geometric feasibility constraints are set as $\bm{w}_{\min} = [-0.06,-0.06,-0.60]^T$ and $\bm{w}_{\max} = [0.06,0.06,-0.40]^T$. 

\subsection{Example 1: Collision avoidance}

We validate the effectiveness of the proposed method in the collision avoidance task. The environment of this example is illustrated in Fig.~\ref{fig_sim_collision_avoidance}. There are two types of obstacles in the environment. The first type of obstacles restrict the motion of the quadcopter base and the size of the flight corridor. The collision avoidance for this type of obstacles is achieved by the flight corridor. The second type of obstacles restrict the motion of the Delta arm and must be avoided through the motion planning of the Delta arm. In Section~\ref{sec_collision_check}, we propose an iterative collision avoidance method to avoid the second type of obstacles. In order to show its effectiveness, the result of the motion planning with the collision avoidance method is calculated.  

Fig.~\ref{fig_sim_collision_avoidance} shows the results of the motion planning with the collision avoidance method and without the collision avoidance method. The generated flight corridor is shown in Fig.~\ref{fig_sim_collision_avoidance}(a). As shown in Fig.~\ref{fig_sim_collision_avoidance}(b), there are four obstacles near the object. In particular, the aerial manipulator collides with one of these obstacles in the resulting trajectory without the collision avoidance method. The collision area is shown as a red dot line in Fig.~\ref{fig_sim_collision_avoidance} and its length is 0.26 m. The resulting trajectory with the collision avoidance method is shown in Fig.~\ref{fig_sim_collision_avoidance} (see the blue line). As can be seen, the result of the proposed method is collision-free. The total computational time for calculating the path, flight corridor, and trajectory of the quadcopter in the collision avoidance task is 46.3 ms, while the computational time for calculating the trajectory of the end-effector is 36.4 ms.

\subsection{Example 2: Aerial retrieval}
The goal of this experiment is to retrieve an object by the aerial manipulator. In the task, the aerial manipulator moves to and picks up the object. Then, the aerial manipulator returns to the start position. The start position is set as $[0, 0, -2.00]$. The position and the orientation angle of the object are set as $[0, -2.00, -1.24]$ and $0^\circ$ (see Fig.~\ref{fig_experiment_aerial_retrieval}(a)), respectively.  As shown in Fig.~\ref{fig_experiment_aerial_retrieval}(b), screens are in the flight environment, which needs to be avoided by the aerial manipulator in the moving stage. The aerial manipulator is set to fly around the screens. As shown in Fig.~\ref{fig_experiment_aerial_retrieval}(a), there are three obstacles near the object. The aerial manipulator has to avoid colliding with these obstacles.

Fig.~\ref{fig_experiment_aerial_retrieval}(b)-(d) shows the result of the aerial retrieval experiment. The generated flight corridor is shown in Fig.~\ref{fig_experiment_aerial_retrieval}(c). The span time of the experiment is 58 s. The mean tracking error of the quadcopter base in the moving stage is 0.05 m, while that in the manipulation stage is 0.01 m. The quadcopter flies faster in the moving stage than in the manipulation stage. However, the higher velocity also causes a larger tracking error. The computational time for calculating the path, flight corridor, and trajectory of the quadcopter in the aerial retrieval task is 43.9 ms, while the computational time for calculating the trajectory of the end-effector is 25.7 ms. 

\subsection{Example 3: Aerial transport}
The goal of this experiment is to grasp an object and put it on the target location. In the task, the aerial manipulator first flies to and picks up the object. Then, the aerial manipulator flies to the target location and puts the object on the target location. Finally, the aerial manipulator returns to the start position. The start position is set as $[0, 0, -2.00]$. The position and the orientation angle of the object are set as $[0, 2.00, -1.22]$ and $0^\circ$, respectively. The position of the target location is set as $[0, -2.00, -1.24]$. The whole process of the experiment is shown in Fig.~\ref{fig_experiment_aerial_pick_place}(c). In the experiment,  the aerial manipulator is also set to fly around the screens to utilize the experiment field. As shown in Fig.~\ref{fig_experiment_aerial_pick_place}(a) and (b), there are two obstacles near the object and three obstacles near the target location. The aerial manipulator has to avoid colliding with these obstacles. 

Fig.~\ref{fig_experiment_aerial_pick_place}(c)-(e) show the result of the aerial transport experiment. The generated flight corridor is shown in Fig.~\ref{fig_experiment_aerial_pick_place}(d). The span time of the experiment is 88 s. The mean tracking error of the quadcopter base in the moving stage is 0.06 m, while that in the manipulation stage is 0.01 m. The computational time for calculating the trajectory of the quadcopter in the aerial transport task is 45.6 ms, while the computational time for calculating the trajectory of the end-effector is 32.8 ms. The experiment result validates the effectiveness of the proposed motion planning method in the aerial transport task.
	
\section{Conclusion}\label{sec_con}
This paper proposed a novel partially decoupled motion planning method of the aerial manipulator for the aerial pick-and-place task. This method calculates the dynamically feasible and collision-free trajectories of the flying base and the manipulator respectively in Cartesian space. The proposed geometric feasibility constraints can ensure the resulting trajectories are coordinated to complete tasks. The proposed method is verified by three experimental results. It is verified by these experiments that the proposed geometric feasibility constraints can ensure the trajectories of the quadcopter base and the end-effector satisfy the geometry of the aerial manipulator. The results also illustrate the ability of the proposed method to avoid obstacles. This ability is limited by the partially decoupled structure since the obstacles nearby the object are avoided by the Delta arm rather than the whole aerial manipulator. However, in order to avoid large obstacles nearby the object, both the quadcopter base and the Delta arm must be used. This will be one important research direction for future research.

\bibliography{myOwnPub,zsyReferenceAll} 
\bibliographystyle{ieeetr}

\end{document}